\documentclass[10pt,twocolumn,letterpaper]{article}

\usepackage{iccv}
\usepackage{times}
\usepackage{epsfig}
\usepackage{graphicx}
\usepackage{amsmath}
\usepackage{amssymb}
\usepackage{multirow}
\usepackage{tabularx}
\usepackage{dsfont}
\usepackage{url}
\usepackage{color}
\usepackage{amsthm}
\usepackage[export]{adjustbox}
\usepackage{comment}

\usepackage[utf8]{inputenc} % allow utf-8 input
\usepackage[T1]{fontenc}    % use 8-bit T1 fonts
\usepackage{url}            % simple URL typesetting
\usepackage{booktabs}       % professional-quality tables
\usepackage{amsfonts}       % blackboard math symbols
\usepackage{nicefrac}       % compact symbols for 1/2, etc.
\usepackage{microtype}      % microtypography
\usepackage{enumitem}
\usepackage{blindtext}% for dummy text only
\usepackage{sidecap}
\usepackage{authblk}
\usepackage{setspace}
\usepackage[pagebackref=true,breaklinks=true,letterpaper=true,colorlinks,bookmarks=false]{hyperref}

\def\iccvPaperID{8729} % *** Enter the ICCV Paper ID here

\definecolor{dg}{rgb}{0,0.694,0.298}
\definecolor{purple}{rgb}{0.4,0.176,0.569}
\definecolor{royalblue}{RGB}{65,105,225}
\usepackage{pifont}% http://ctan.org/pkg/pifont
\usepackage[breaklinks=true,bookmarks=false]{hyperref}

\iccvfinalcopy % *** Uncomment this line for the final submission

\def\iccvPaperID{****} % *** Enter the ICCV Paper ID here

\ificcvfinal\fi

\begin{document}

%%%%%%%%% TITLE
\title{CDUL: CLIP-Driven Unsupervised Learning for Multi-Label Image Classification}
\vspace{-0.9 cm}
\author[1]{Rabab Abdelfattah}
\author[2]{Qing Guo}
\author[3]{Xiaoguang Li}
\author[3]{Xiaofeng Wang}
\author[3]{Song Wang}

\affil[1]{University of Southern Mississippi, USA
\authorcr
 \tt rabab.abdelfattah@usm.edu}
 
\affil[2]{IHPC and CFAR, Agency for Science, Technology and Research, Singapore  
\authorcr
 \tt tsingqguo@ieee.org}
 
\affil[3]{University of South Carolina, USA \authorcr
  \tt xl22@email.sc.edu, \{\tt wangxi, songwang\}@cec.sc.edu}
  % }
\vspace{-0.9 cm}

\maketitle
\ificcvfinal\thispagestyle{empty}\fi

\begin{abstract}
\vspace{-0.3 cm}
This paper presents a CLIP-based unsupervised learning method for annotation-free multi-label image classification, including three stages: initialization, training, and inference.  At the initialization stage, we take full advantage of the powerful CLIP model and propose a novel approach to extend CLIP for multi-label predictions based on global-local image-text similarity aggregation.  To be more specific, we split each image into snippets and leverage CLIP to generate the similarity vector for the whole image (global) as well as each snippet (local). Then a similarity aggregator is introduced to leverage the global and local similarity vectors. Using the aggregated similarity scores as the initial pseudo labels at the training stage, we propose an optimization framework to train the parameters of the classification network and refine pseudo labels for unobserved labels.  During inference, only the classification network is used to predict the labels of the input image. Extensive experiments show that our method outperforms state-of-the-art unsupervised methods on MS-COCO, PASCAL VOC 2007, PASCAL VOC 2012, and NUS datasets and even achieves comparable results to weakly supervised classification methods.
\end{abstract}

%%%%%%%%% BODY TEXT

\vspace{-0.5 cm}
\section{Introduction}
\begin{figure}[!t]
  \centering
  	\includegraphics[scale=0.37]{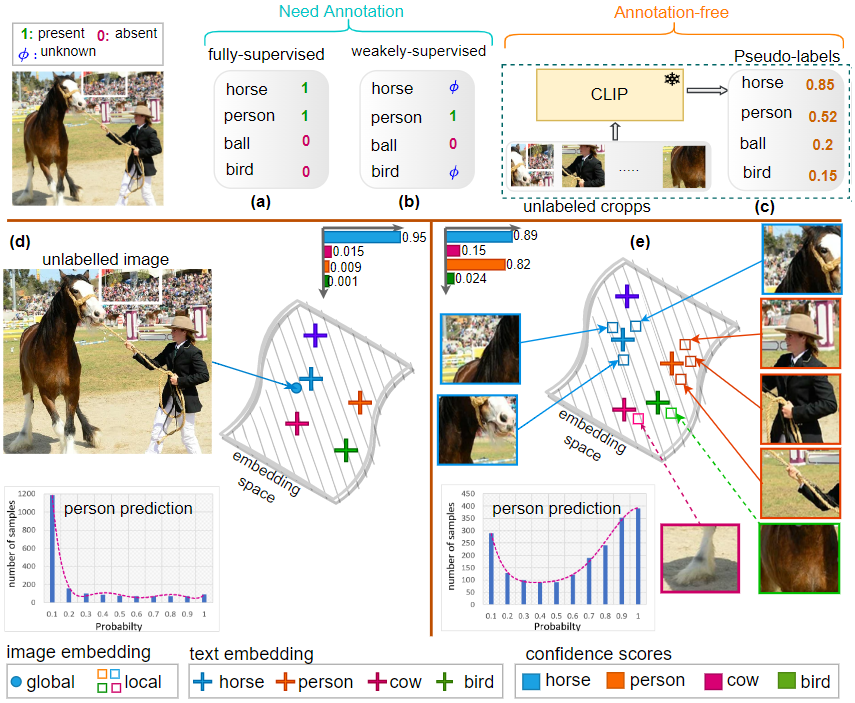}\\
  % \vspace{-0.25 cm}
  \caption{A comparison of our solution with fully and weakly-supervised multi-label classification. (a) The training dataset images for fully-supervised learning are fully labeled. (b) The training images used in weakly-supervised are partially labeled. (c) Our unsupervised multi-label classification method is annotation-free. (d) CLIP focuses on one class in the whole image, and the embedding is denoted by blue circle. Some classes are ignored such as "person". (e) In our approach, image snippets are mapped separately to the embedded space, where each snippet's embedding is denoted by squares. Local alignment allows to predict more labels. }
  \label{fig:main-fig}
   \vspace{-0.55 cm}
\end{figure}

%\vspace{-0.1 cm}
% Multi-label definition and challanges
A multi-label classification task aims to predict all the objects within the input image, which is advantageous for various applications, including content-based image retrieval and recommendation systems, surveillance systems, and assistive robots, to name a few~\cite{chen2019multi,chen2019learning,chen2020knowledge}. However, getting clean and complete multi-label annotations is very challenging and not scalable, especially for large-scale datasets, because an image usually contains multiple labels (Figure~\ref{fig:main-fig}.a). 
%
%%%%%%%%%%%%%%%%%%%%%%%%%%%%%%%%%%%%%%%%%%
 
%############################################
%weakly supervised models
To alleviate the annotation burden, weakly supervised learning approaches have been studied~\cite{durand2019learning,huynh2020interactive,cole2021multi,abdelfattah2022g2netpl}, in which only a limited number of objects are labeled on a subset of training images (Figure~\ref{fig:main-fig}.b).  Though less than the fully-labeled case, it still requires intensive manpower and time for annotations.

To go one step further, we consider unsupervised multi-label image classification, leveraging the off-the-shelf vision-language models such as contrastive language-image pre-training (CLIP)~\cite{radford2021learning}. %to generate the pseudo labels for unlabeled data for training the models.
CLIP is trained by matching each input image to the most relevant text description over 400 million image-text pairs collected from the Internet. It has demonstrated remarkable zero-shot classification performance as a pre-trained model in image-text retrieval~\cite{radford2021learning}, video-text retrieval~\cite{luo2022clip4clip}, and
single-label image classification~\cite{radford2021learning}. With CLIP, the encoded visual representations can be directly used for vocabulary categorization without additional training.  However, CLIP is not suitable for multi-label classification, since it is trained only for recognizing a single object per image (Figure \ref{fig:main-fig}.d).
Finding only one global embedding for the whole image may push CLIP to generate a high confidence score for the closest semantic text class, while neglecting other classes. In Figure~\ref{fig:main-fig}.d, for instance, CLIP predicts class \textit{"horse"} with a very high confidence score (0.98), but gives a very low weight to class \textit{"person"}, given the fact that CLIP suffers from excessive polysemy~\cite{radford2021learning}. 
%
%%%%%%%%%%%%%%%%%%%%%%%%%%%%%%%%%%%%%%%%%%
% \vspace{-0.5 cm}
\begin{figure*}[!t]
  \centering
  	\includegraphics[scale=0.38]{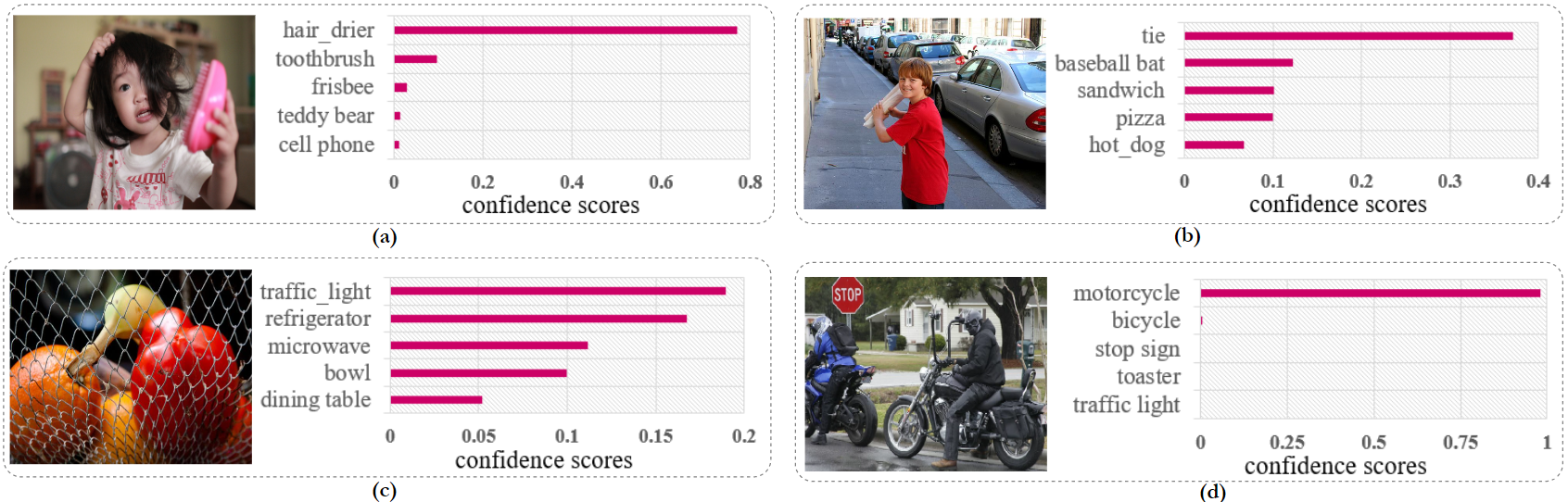}\\
  \vspace{-0.1 cm}
  \caption{Confidence scores from the off-the-shelf CLIP on sample images from COCO dataset}
  \label{fig:clip-pred}
  \vspace{-0.3 cm}
\end{figure*}
%############################################

To address these issues and make full use of CLIP in multi-label classification, this paper presents a CLIP-driven unsupervised learning method \textcolor{black}{(CDUL)} for multi-label image classification, which includes three stages: initialization, training, and inference. At the initialization stage, we use CLIP to generate global representation of the whole image and, more importantly, local representations of snippets of the image. A novel aggregation of global and local representations provides high confidence scores for objects on the image.  As shown in Figure~\ref{fig:main-fig}.e, the class ``person'' receives high confidence score in this case.  At the training stage, the confidence scores will be used as the initial values of pseudo labels, with which a self-training procedure is proposed to optimize the parameters of the classification network as well as the pseudo labels.  Finally, during inference, only the classification network is used to predict the labels of an~image.
% \textcolor{blue}{From a learning task perspective, our method can be classified as unsupervised learning: \ding{182} Our method does not require ground truth labels. In the field of multi-label classification, weakly supervised and fully supervised methods are categorized based on the ratio of ground-truth labels they can access. \ding{183} The pseudo labels generated by CLIP are not treated as ground-truth labels directly and we design a new method to refine them during training. 
% \ding{184} Our method shares similarities with self-supervised learning where the networks are explicitly trained with automatically generated labels and self-supervised learning is a subset of unsupervised learning methods \cite{jing2020self}. }
%
%In summary, we propose a novel method that can effectively optimize the pseudo labels based on the local-global representations of input images for a annotation-free multi-label image classification task. Specifically, this paper makes three-fold contributions:

The contributions of this paper are listed as follows:
    \vspace{-1mm}
\begin{itemize}
    \item 
    We propose a novel method for unsupervised multi-label classification training. To the best of our knowledge, this is the first work that applies CLIP %to generate pseudo labels 
    for unsupervised multi-label image classification.  The aggregation of global and local alignments generated by CLIP can effectively reflect the multi-label nature of an image, which breaks the impression that CLIP can only be used in single-label classification. 
    \vspace{-2mm}
    % To the best of our knowledge, this is the first work that uses off-the-shelf CLIP to generate the pseudo labels for training unsupervised multi-label classification models without training or tuning CLIP. 
%    \item 
%     We propose an analysis for the CLIP prediction based on the global alignment using the whole image and the local alignment using image snipping. Moreover, we present a new aggregator to combine the global-local similarity vectors to produce high-quality pseudo labels. 
    \item A gradient-alignment training method is presented, which recursively updates the network parameters and the pseudo labels.  By this algorithm, the classifier can be trained to minimize the loss function.
        \vspace{-2mm}
    % Furthermore, we proposed an optimization framework that learns based on alternatively updating both classification network parameters and the pseudo labels latent parameters to improve the classification task further.
    \item Extensive experiments show that our method not only outperforms the state-of-the-art unsupervised learning methods, but also achieves comparable performance to weakly supervised learning approaches on four different multi-label datasets.
    % , but 
\end{itemize}

\vspace{-1.5mm}
% The rest of the paper is organized as follows.  Section~\ref{sec:rw} discusses the related work.  Section~\ref{sec:me} proposes the methodology.  Experimental results are presented in Section~\ref{sec:exp} and Conclusions are drawn in Section~\ref{sec:con}.

\section{Related Work}
\label{sec:rw}

\vspace{-1.5mm}
\noindent
{\bf Weakly Supervised Multi-Label Classification.}
Due to high annotation costs, weakly supervised learning in multi-label classification becomes an interesting topic of research. Weakly supervised models are trained on a partial-label setting where some labels are annotated (called ``observed labels''), and the rest are not annotated (called ``unobserved or unknown labels''). Early work includes assuming the unobserved labels as negative~\cite{bucak2011multi,sun2010multi,wang2014binary},predicting the unobserved labels using label correlation modeling~\cite{deng2014scalable,wu2015ml,xu2013speedup}, and probabilistic modeling~\cite{vasisht2014active,kapoor2012multilabel}. However, these approaches rely on traditional optimization and cannot be scaled to train deep neural networks (DNNs). 
Recently, research effort has been made to train DNNs using partial labels~\cite{durand2019learning,huynh2020interactive,cole2021multi,Pu2022SARB,chen2022structured,sun2022dualcoop,kim2022large}. In general, these approaches can be divided into two groups. The first group uses observed labels as the ground truth to build the label-to-label similarity graph \cite{huynh2020interactive}, cross-images semantic correlation \cite{chen2022structured}, encodes positive and negative contexts with class names \cite{sun2022dualcoop}, and blend category-specific representation across different images \cite{Pu2022SARB}. The second group starts with a subset of observed labels and soft pseudo labels for unobserved labels, and update pseudo labels during training, such as \cite{szegedy2016rethinking,mac2019presence,cole2021multi,durand2019learning}. Different from all these models, our method works without the need of annotations. %Therefore, for a fair comparison, we compare the former group under weakly supervised methods by proving some observed labels to keep their performance and compare the latter group as unsupervised methods by initializing all the weakly supervised models, including our model, with our proposed soft pseudo labels for unlabeled samples during the training of the methods.

\noindent
{\bf Vision-Language Pre-Training.}
Vision-language pre-training models achieve impressive performance on various tasks. Several techniques for learning visual representations from text representations have been presented using semantic supervision~\cite{miech2020end,radford2021learning,wang2021simvlm}. Among these models, the most effective one is CLIP~\cite{radford2021learning}, which exploits the large-scale image-text pairs collected from the Internet to achieve alignment of images and text representations in the embedding space. CLIP leverages contrastive learning, high-capacity language models, and visual feature encoders to efficiently capture interesting visual concepts. It shows remarkable performance in different tasks such as zero-shot inference and transfers learning in single image classification~\cite{jia2021scaling,radford2021learning}. However, CLIP is trained to focus on global representation, since the input image and text description both contain global semantic information. As a result, it only predicts the closest semantic text class, while neglecting other classes.  Our method takes a different route by proposing a model to learn both global and local visual representations to enrich semantic concepts in multi-label classification. 
{There are approaches using the CLIP model at the pre-training stage to help the models first develop a general understanding of the relationship between visual and textual concepts, such as RegionCLIP for object detection~\cite{zhong2022regionclip}. However, to fine-tune these pre-trained models, a large amount of labeled data is still needed, which does not belong to the category of weakly or unsupervised learning.}
%which can be expensive and time-consuming, particularly for large datasets. The model is first pre-trained on a large corpus of text and images using the CLIP pre-training process, which helps the model develop a general understanding of the relationship between visual and textual concepts.  Then the model is fine-tuned using labeled object detection data that usually consists of images and their corresponding object annotations.}
%

\noindent
{\bf Unsupervised Feature Learning.}
% Some methods have been proposed for unsupervised multi-label classification in person re-identification such as \cite{wang2020unsupervised,zhang2022implicit}. However, these methods focus on learning the identity features for person ReID which is not related to our topic.
% Recently, self-supervised learning approaches \cite{chen2020simple,he2020momentum,hjelm2018learning,wu2018unsupervised,zhuang2019local} are adopted for unsupervised pre-training tasks where a contrastive loss is used to acquire instance-discriminative representations. However, these approaches require networks to be fine-tuned using the ground-truth labels on downstream tasks, which does not fit our unsupervised multi-label classification problem~\cite{zhang2021refining}. Another direction is to customize weakly supervised learning algorithms for unsupervised learning.  In fact, the pseudo-label-based weakly supervised algorithms~\cite{szegedy2016rethinking,mac2019presence,cole2021multi,durand2019learning} can be easily modified for unsupervised multi-label classification by assigning pseudo labels to all objects without using the observed labels. In experiments, we will compare our solution to these methods.
Some methods for unsupervised multi-label classification in person re-identification \cite{wang2020unsupervised,zhang2022implicit} focus on identity features, which are not relevant to our topic. Self-supervised learning approaches \cite{chen2020simple,he2020momentum,hjelm2018learning,wu2018unsupervised,zhuang2019local} use contrastive loss for instance-discriminative representations, but require ground-truth labels for fine-tuning, which does not suit our unsupervised multi-label classification problem \cite{zhang2021refining}. Pseudo-label-based weakly supervised algorithms \cite{szegedy2016rethinking,mac2019presence,cole2021multi,durand2019learning} can be easily adapted for unsupervised multi-label classification by assigning pseudo labels to all objects without using observed labels. We will compare our solution to these methods in experiments.

\section{Methodology}
\label{sec:me}
\noindent
\textbf{Notations.} Let $\mathcal{X} = \{x_1, x_2,...x_M\}$ denote the training set, where $M$ is the number of images in $\mathcal{X}$ and $x_m$ for $m=1\cdots,M$ is the $m$th image.  In our formulation, $\mathcal{X}$ is totally unlabeled. Let $C$ be the total number of classes in the dataset. Let $y_{u,m} \in \mathbb R^{C}$ denote the pseudo label vector of image $x_m$.  Notice that each entry of $y_{u,m}$ belongs to the interval $[0,1]$.  The overall pseudo label set is denoted by $Y_u = [y_{u,1},y_{u,2},....,y_{u,M}] \in \mathbb R^{C \times M}$. We also define the latent parameter vector of $\tilde y_{u,m}$ as $\tilde y_{u,m} = \sigma^{-1}(y_{u,m}) \in \mathbb R^{C}$  for image $x_m$ where $\sigma$ is the sigmoid function.  The prediction set is $Y_p = [y_{p,1},y_{p,2},....,y_{p,M}]$ where $y_{p,m} \in \mathbb R^{C}$ is the vector of the predicted labels for image $x_m$. In CLIP model, there are two encoders: the visual encoder and the text encoder, which are denoted by $E_v$ and $E_t$, respectively. The visual encoder maps the input image $x_m$ to the visual embedding vector $E_v(x_m) = f_m \in \mathbb{R}^{K}$ where $K$ is the dimension length of the embedding. Similarly, the text encoder maps the input text (class $i$, $i=1,\cdots,C$) to the text embedding vector $E_t(i) = w_i \in \mathbb{R}^{K}$. Here the input text is a predefined prompt, such as ``a photo of a cat''. Given a vector or a matrix $Q$, $Q^\top$ means the transpose of $Q$.

\begin{figure*}[!h]
  \centering
  	\includegraphics[scale=0.44]{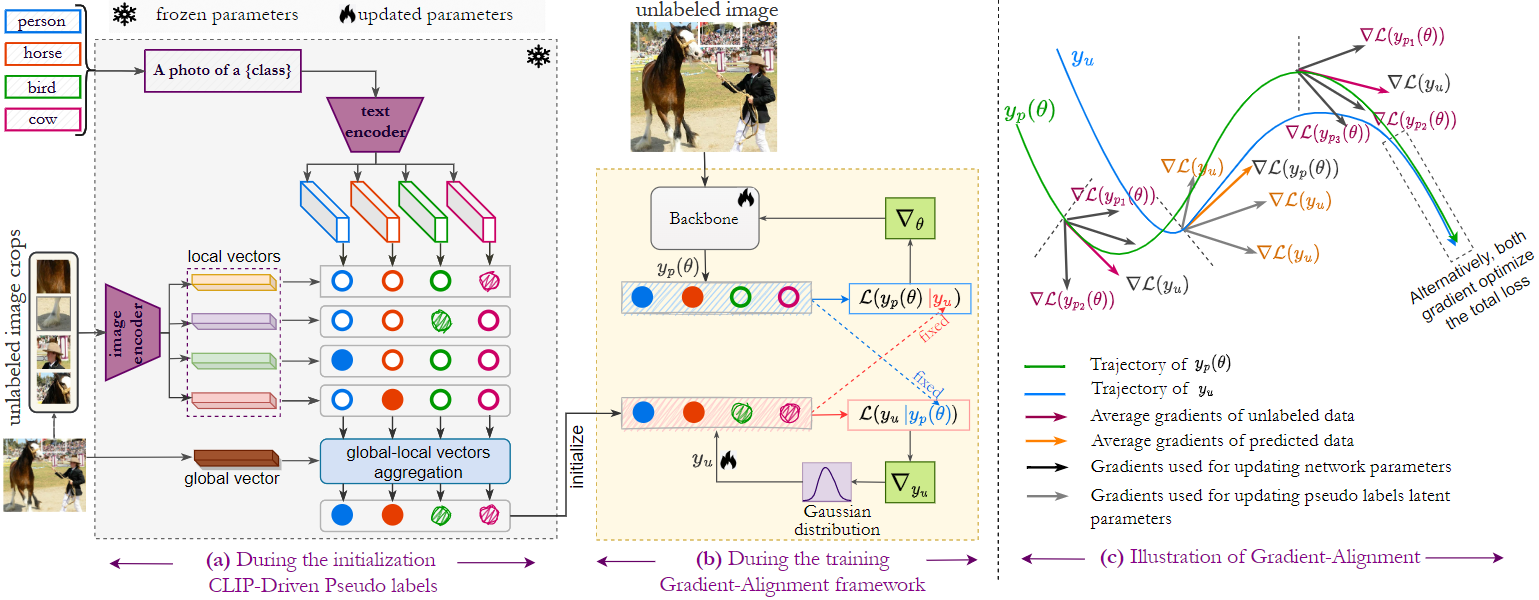}\\
  % \vspace{-0.05 cm}
  \caption{The overall framework for CDUL unsupervised multi-label image classification. \textbf{(a) During initialization}, we propose CLIP-driven global and local alignment and aggregation to generate pseudo labels. ($i$) Given an image, CLIP predicts the global similarity vector $S^{global}$; ($ii$) Given the snippets of this image, CLIP predicts local similarity vectors $S_j^{local}$; ($iii$) The global-local aggregator is used to generate the pseudo labels $S^{final}$. \textbf{(b) During training}, the pseudo labels generated from initialization are use to supervise the training of the classification network, using our proposed method \textit{gradient-alignment method}. \textbf{(c)} The \textbf{gradient alignment illustration} shows that updating the network parameters and the pseudo labels by turns pushes both the pseudo label $y_u$ and the predicted label $y_p$ to the optimal solution to minimize the total loss function. \textbf{During inference}, we apply the whole image to the classification network to get the multi-label predictions.}
  \label{fig:clip-main-framework}
  \vspace{-0.3 cm}
\end{figure*}

\smallskip
\noindent
\textbf{Overview.} The proposed framework is shown in Figure \ref{fig:clip-main-framework}, to address unsupervised multi-label image classification, which includes three stages:  initialization, training, and inference. During initialization, the goal is to appropriately initialize the pseudo labels for the unobserved labels on each training image.  Taking advantage of the off-the-shelf CLIP model, we propose a CLIP-driven approach to build the pseudo labels upon the aggregation of global and local semantic-visual alignments, which can significantly improve the quality of pseudo-labels. 
During training, the pseudo labels obtained in initialization will be used as the estimation of the unobserved labels to initialize training of the classification network.  We propose an optimization method that minimizes the total loss by recursively updating the network parameters and the latent parameters of the pseudo-labels. During inference, only the classification network is used to predict the labels of the input image. 

\subsection{Pseudo Label Initialization}
\subsubsection{Global Alignment Based on CLIP}

CLIP is a powerful vision-language model that focuses on learning the global representation (the dominant concept) of an image (Figure~\ref{fig:clip-pred}). Therefore, we can directly use the CLIP model to generate the global alignment of an image without tuning the model parameters.  Since the following discussion only focuses on an individual image $x_m$, we will drop the index $m$ for notational simplicity.

Given an input image, the visual encoder of CLIP maps it to the embedding vector $f$. The relevant similarity score between $f$ and the text embedding $w_i$ is given by
\begin{eqnarray}
p_{i}^{glob} & = & \frac{f^{\top} w_i}{||f|| \cdot ||w_i||},~~ \forall 1 \leq i \leq C
\label{eq:cosim}
\\
s_{i}^{glob} & = & \frac{{\rm exp}(p_{i}^{glob}/\tau)}{\sum_{i=1}^{C} {\rm exp}(p_{i}^{glob}/\tau)}, \label{eq:cosim1}
\vspace{-0.5 cm}
\end{eqnarray}
where $p_{i}^{glob}$ denotes cosine similarity score between $f$ and $w_i$ for class $i$ on the input image, $s_{i}^{glob}$ is the normalized value for the similarity score using softmax, and $\tau$ is the temperature parameter learned by CLIP. Then, the soft global vector of this image is defined as \[S^{global} = \{s^{glob}_{1},s^{glob}_{2},....,s^{glob}_{C}\},\] which includes the similarity score for each class. 
% As a result of this low temperature, one class has a significantly higher probability among the other classes. 

%%%%%%%%%%%%%%%%%%%%%%%%%%%%%%%%%%%%%%%%%%
% \vspace{-0.5 cm}

%#######################################
Notice that CLIP focuses on the most relevant class on an image and is limited to predict only one single label per image, while an image may contain multiple labels. In some cases, the highest confidence score from the CLIP prediction is not even correct due to the lack of appropriate prompt design (Figure~\ref{fig:clip-pred}.a). To alleviate this issue, we shift CLIP's attention from global to the local level, i.e., using CLIP to predict snippets of an image rather than the entire image, which will be discussed in the next subsection.
\smallskip
\noindent
\subsubsection{CLIP-Driven Local Alignment}
To generate the local alignment, we split an input image to $N$ snippets, denoted by $\{r_j\}_{j=1,...,N}$. Each snippet may contain multiple objects rather than just a single object.
Accordingly, the visual embedding vector $g_j \in \mathbb R^K$ of snip $r_j$ is extracted from the visual encoder, $E_v(r_j) = g_j$. 
% which results in visual snip embedding $v = \{v_1, v_2, ..., v_{n-1}\} \in \mathbb{R}^{n \times k}$. 
Each image snippet is handled separately by finding the cosine similarity scores $p_{j,i}^{loc}$ between the snippet visual embedding $g_j$ and the text embedding $w_i$ for class $i$:
\vspace{-0.25 cm}
\begin{equation}
p_{j,i}^{loc} = \frac{g_j^{\top} w_i}{||g_j|| \cdot ||w_i||},~~ \forall~ 1 \leq j \leq N,~1 \leq i \leq C
\label{eq:cosim2}
\vspace{-0.15 cm}
\end{equation}
The similarity scores will be forwarded to the Softmax function that normalizes these scores over all classes:
\vspace{-0.15 cm}
\begin{equation} 
s_{j,i}^{loc} =  \frac{{\rm exp}(p_{j,i}^{loc}/\tau)}{\sum_{i=1}^{C} {\rm exp}(p_{j,i}^{loc}/\tau)}.
\label{eq:cosim4}
\vspace{-0.15 cm}
\end{equation}
So the local soft similarity vector $S_j^{local}$ of snippet $r_j$ is given~by
\vspace{-0.25 cm}
\[
S_{j}^{local} = \{s^{loc}_1,s^{loc}_2,....,s^{loc}_C\}.
\]
% $n$ represent the image subregion's vectors, and the last one-hot vector implements the whole input image without cropping. To extract pseudo labels,
% the category prediction at each subregion and whole image is an argmax of classification soft similarities scores along the class dimension. The intuition behind using the maximum is that the score of classification should depend on the highest probability assigned to any known classes.

%%%%%%%%%%%%%%%%%%%%%%%%%%%%%%%%%%%%%%%%%%
% \vspace{-0.5 cm}
\begin{figure*}[!t]
  \centering
  	\includegraphics[scale=0.37]{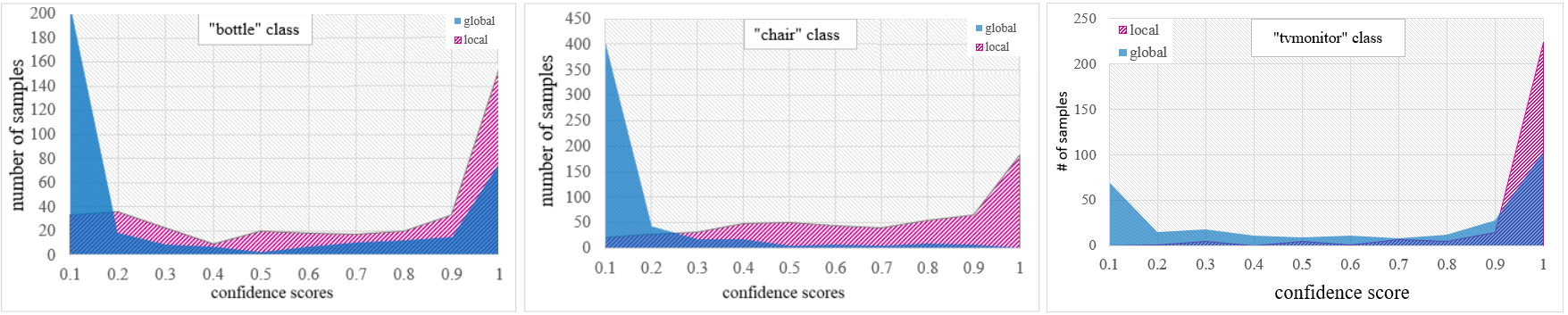}\\
  % \vspace{-0.25 cm}
  \caption{The distributions of the predicted labels across the confidence scores using off-the-shelf CLIP on the whole image (global) and snappets (local). }
  \label{fig:histogram}
   \vspace{-0.4 cm}
\end{figure*}
%############################################

Notice that different snippets may contain different objects or different attributes for the same object. Therefore, a specific class, which cannot obtain the highest similarity score from CLIP when focusing on the entire image, may now get the highest score in several snippets. Such a local alignment can enhance semantic transfer per snippet.
% The first stage generates pseudo-labels of entity categories given a cropped image without dense annotations other than web-sourced image-text pairings with potentially noisy alignment. 
Figure~\ref{fig:histogram} shows the comparison of the confidence score distributions of using CLIP to predict three classes (``bottle'', ``chair'', and ``tvmonitor'') in PASCAL VOC 2012 dataset, using global images and local snippets, respectively. 
%%%%%%%%%
It can be observed that when focusing on global images, CLIP may neglect some classes due to the ``domain gap'' between the pre-training datasets used to train CLIP and the target multi-label dataset.  For instance, in Figure~\ref{fig:histogram}.b, the ``chair'' class get very low scores in most images, which means that very few ``chair'' labels will be predicted.  This will affect the training performance at the training stage.  When snippets are considered, they can enhance the prediction distribution toward higher confidence~scores.  {It is worth mentioning that, as a cropping method, CLIP-Driven Local Alignment (CDLA) has advantages over class-agnostic object detection (COD)~\cite{jaiswal2021class}. Our CDLA does not need the ground truth to get the snippets, while COD needs the ground truth to train the model to extract the snippets containing the objects. Thus, our CDLA is less expensive than COD in computation. Moreover, our CDLA technique is more robust than COD in the situations where the object of interest is occluded or partially visible. In those cases, object detection methods may not be able to detect the object, while cutting images may still be valid to obtain a partial view of the object.}
%
%since the global semantic of the image can be easily broken by applying image augmentations such as cropping the image, which is the reason for finding new objects based on the alignment of the semantic of each snip with the text \cite{lee2022uniclip}. 
%

%(3) CLIP aligns the semantic image vector with the closest text embedding vector. For example, the "\textit{horse}" class is most likely one of the seen classes, so it is easy for CLIP to predict it without cropping. In the case of the "\textit{chair}" class, maybe multiple fine-grained class names are used, such as "\textit{seat}, \textit{bench}, \textit{stool}, \textit{settee}, \textit{couch}" which are most related to the global semantics of the image. 
\vspace{-0.5 cm}
\subsubsection{Global-Local Image-Text Similarity Aggregator}
Note that each input image is associated with one global image-text similarity vector $S^{global}$ and $N$ local similarity vectors $S_j^{local}$. We propose an aggregation strategy that complements mutual information from $S_j^{local}$ and generates a unified local similarity vector $S^{aggregate}$ for each image, using a min-max method.
\vspace{-0.25 cm}
 Let
\begin{align*} \centering
\alpha_{i} & = \max_{j=1,\cdots, N}s_{j,i}^{loc}, \\
\beta_{i} & = \min_{j=1,\cdots, N} s_{j,i}^{loc},~~~  \forall~ 1 \leq i \leq C
\end{align*}
and
\vspace{-0.25 cm}
\begin{equation}
\gamma_i = \left\{
    \begin{array}{ll}
        1 & \alpha_i \ge \zeta\\
        0 & \alpha_i < \zeta
    \end{array}
\right.
\vspace{-0.15 cm}
\end{equation}
where $\zeta$ is the threshold parameter.  The aggregation score for class $i$ is given by
\vspace{-0.25 cm}
\begin{equation} \centering
s_i^{ag} = \gamma_i \alpha_i + (1-\gamma_i) \beta_i.
\vspace{-0.25 cm}
\end{equation}
This strategy basically means that if the highest similarity score that class $i$ obtains among all snippets, $\alpha_i$, is greater than $\zeta$, we will consider that this class likely exists on the image with $\alpha_i$ assigned to $s_i^{ag}$.  On the contrary, if the similarity scores of class $i$ in all snippets are less than $\zeta$, the likelihood of class $i$ existing on this image is small.  Therefore, the strategy assigns the minimum score $\beta_i$ to $s_i^{ag}$.  With the aggregation scores, we define the soft aggregation vector of all classes for each input image as follows:
\vspace{-0.25 cm}
\[
S^{aggregate} = \{s^{ag}_1,s^{ag}_2,....,s^{ag}_C\} .
\vspace{-0.25 cm}
\]

Now we can leverage the global similarity, which adds more comprehensive and complementary semantics, to local similarity by calculating the average:
\vspace{-0.25 cm}
\begin{equation*}
S^{final} = \frac{1}{2} \left(S^{global} + S^{aggregate}\right),
\vspace{-0.25 cm}
\end{equation*}
which will be used as the initial pseudo labels for unobserved labels at the training stage.  The high quality of $S^{final}$ will significantly enhance the training performance, which is discussed in Subsection~\ref{subsec:Quality}.
%  %%%%%%%%%%%%%%%%%%%%%%%%%%%%%%%%%%%%%%%%%%%%%%%%%%%%%%%%%%%%%%%
\begin{table*}[!t]
    \centering{
    \caption{Mean average precision mAP in (\%) for different multi-label classification methods under different supervision levels: Fully supervised, Weakly supervised and unsupervised, in addition to compare to zero-shot CLIP for four different datasets. Blue color represents the best results.}
    \label{tab:supervised-level}
\begin{adjustbox}{width=1 \textwidth}
    \begin{tabular}%{|l|l|l|l|l|}
        {
    %   |
      >{}p{0.2\textwidth}|
      >{}p{0.16\textwidth}|
     >{}p{0.16\textwidth}|
      >{\centering}p{0.13\textwidth}|
     >{\centering}p{0.13\textwidth}|
      >{\centering}p{0.13\textwidth}|
      >{\centering\arraybackslash}p{0.13\textwidth}
    %   |
    }
\hline

        Supervision level & Annotation & Method   & VOC2012 & VOC2007 & COCO & NUS  \\ \hline \hline
                %%%%%%%%%%%%%%%%%%%%%%%%%%%%%%%%%%%%
         % Zero-shot &- &  CLIP  \cite{radford2021learning} & 85.3 &86.2  & 65.3 & ~ \\ \hline\hline
        %%%%%%%%%%%%%%%%%%%%%%%%%%%%%%%%%%%%%
        \multirow{2}{*}{Fully Supervised}  & \multirow{2}{*}{Fully labeled}  
        %bce
        & BCE-LS ~\cite{cole2021multi}&91.6 &92.6 &79.4 &51.7 \\ \cline{3-7}
        %bce
        ~& ~& BCE  &90.1 &91.3 &78.5 & 50.7  \\ \hline \hline
        %%%%%%%%%%%%%%%%%%%%%%%%%%%%%%%%%%%
         \multirow{5}{*}{Weakly Supervised}  & \multirow{3}{*}{10\% labeled} & SARB \textit{et al.}\cite{Pu2022SARB} & - &85.7 & 72.5 & -  \\ \cline{3-7}
        %AAAI (10% partial labels resnet-101)
        ~ & ~  &ASL \textit{et al.}\cite{ben2020asymmetric} & - &82.9 & 69.7 & -  \\ \cline{3-7}
        ~ & ~  &Chen \textit{et al.}\cite{chen2022structured} & - &81.5 & 68.1 & -  \\ \cline{2-7}    
        ~ & \multirow{2}{*}{one observed labeled}  &  LL-R \cite{kim2022large} & 89.7 & 90.6 & 72.6 & 47.4 \\ \cline{3-7}
        ~ & ~ &  G$^2$NetPL \cite{abdelfattah2022g2netpl} & 89.5 &89.9  & 72.5 & 48.5 \\ \hline \hline

         \multirow{6}{*}{Unsupervised} & \multirow{6}{*}{Annotation-free} & Naive AN \cite{kundu2020exploiting} & 85.5 &86.5 & 65.1 & 40.8 \\ \cline{3-7}
         %ANLS
        & ~ & Szegedy \textit{et al.}~\cite{szegedy2016rethinking} & 86.8 &87.9 & 65.5 & 41.3 \\ \cline{3-7}
        %WAN
         & ~ & Aodha \textit{et al.}\cite{mac2019presence} & 84.2 &86.2 & 63.9 & 40.1 \\ \cline{3-7}
         %curriculum learning
         & ~ &   Durand \textit{et al.}\cite{durand2019learning} & 81.3 &83.1 & 63.2 & 39.4 \\ \cline{3-7}
          % & ~ & Huynh \textit{et al.}\cite{huynh2020interactive} & ~ & ~ & ~ \\ \cline{3-5}
          %epr
         % ~ & ~ & Cole \textit{et al.}\cite{cole2021multi} & ~ & ~ & ~ \\ \cline{3-6}
         %ROLE
         ~ & ~ & ROLE~\cite{cole2021multi} & 82.6 &84.6 & 67.1 & 43.2  \\ \cline{3-7}

         ~ & ~ &  \textcolor{blue}{CDUL (ours)} & \textcolor{blue}{88.6} & \textcolor{blue}{89.0} & \textcolor{blue}{69.2} & \textcolor{blue}{44.0} \\ \cline{1-7}
\hline
% \hline
    \end{tabular}
        \end{adjustbox}
    }
    \vspace{-0.5 cm}
\end{table*}
%%%%%%%%%%%%%%%%%%%%%%%%%%%%%%%%%%%%%%%%%%%%%%%%%%%%%%
\subsection{Gradient-Alignment Network Training}
\label{subsec:GANT}
This subsection proposes the gradient-alignment method to leverage unsupervised consistency regularization, which updates the network parameters and the pseudo labels by turns. To be more specific, one can first train the network parameters according to the Kullback-Leibler (KL) loss function between the predicted labels and the initial pseudo labels obtained from the initialization stage, treating the pseudo labels as constants.  After that, we fix the predicted labels and employ the gradient of the loss function with respect to the pseudo labels to update the latent parameters of pseudo labels.  Once the pseudo labels are updated, we can fix them again and re-update the network parameters.  This optimization procedure will continue until convergence occurs or the maximum number of epochs is reached.
%
% \subsubsection{Updating Pseudo Labels Gradient-Alignment}
This idea is inspired by the previous work~\cite{zhang2021refining,cho2022part,abdelfattah2022plmcl}, which shows that during the training process the previously generated pseudo labels can provide valuable information to supervise the network. 
The detailed procedure can be described as follows.  At the beginning of training, the pseudo label vector is initialized by $S^{final}$ from the global-local aggregation module, i.e., $y_u = S^{final}$.  
%Then the latent parameters of the pseudo labels can are
%\begin{align}
%\tilde y_{u} = \sigma^{-1}(S^{final})
%\end{align}
%It is worth noting that we start the training without annotated labels, the estimated labels from CLIP are employed to supervise the classification network. As a result, there is no clear signal for the right and wrong labels. Therefore, 
Then $y_u$ will be fixed and used to supervise the train of the network with the Kullback-Leibler (KL) loss function
$\mathcal L(Y_p | Y_u, \mathcal X)$.  When the training is done, we fix the predicted labels $Y_p$ and update the latent parameters of pseudo labels $\tilde y_u$:
\vspace{-0.25 cm}
\begin{equation}
\tilde y_{u}= \tilde y_{u} - \psi (y_{u})\circ  \nabla_{y_u} \mathcal{L} (Y_{u} | Y_p, \mathcal X)
\label{eq:update_vector}
\vspace{-0.25 cm}
\end{equation} 
where $\circ$ means the element-wise multiplication, $y_u = \sigma(\tilde y_u)$, and $\psi (y_{u})$ is a Gaussian distribution with the mean at 0.5 which is given by:
\vspace{-0.25 cm}
\begin{align}
    \psi ([{y}_{u}]_i) = \frac{1}{\sigma \sqrt{2\pi}} e^{- \frac{1}{2} \left(\frac{[{y}_{u}]_i-0.5}{\sigma}\right)^2}.
    \label{eq:psi}
    \vspace{-0.25 cm}
\end{align}
Here $[{y}_{u}]_i$ is the $i$th entry of the vector $y_u$. 
Since the whole dataset is unlabelled, we need to use $\psi(y_{u})$ to increase the rate of change for the unconfident pseudo labels and reduce the rate for the confident pseudo labels. The Gaussian distribution can perform as such a function for pseudo labels. The confidence of the pseudo label is evaluated based on $2|[y_u]_i-1|$. For instance, if $[y_{u}]_i$ is 0.5, the Gaussian distribution will achieve its maximal value, which means that our module is not confident about the pseudo label and the rate of change should contribute more in the iteration of the pseudo label to push it away from 0.5.  Otherwise, if $[y_{u,t}]_i = 0$ or $[y_{u,t}]_i = 1$, 
the Gaussian distribution value reaches its minimum, which indicates high confidence on the current pseudo label and the rate of change should contribute less so that the value of the pseudo label can be maximally kept. 

Once $Y_u$ is updated, we switch back to the network training with a fixed $Y_u$ again.  This procedure will continue until convergence takes place or the maximum number of epochs is reached.
As shown in Figure~\ref{fig:clip-main-framework}.c, this training process pushes both $Y_u$ and $Y_p$ (the predictions is a function of the network parameters) to a non-trivial optimal point to minimizing $\mathcal L(Y_p, Y_u | \mathcal X)$. 

\subsection{Inference}  
\vspace{-0.2 cm}
We simply feed the whole image, without splitting, to the classification network to get the prediction. It is worth noting that we use the whole image without cropping during the training and testing process to reduce the computational~cost.

%%%%%%%%%%%%%%%%%%%%%%%%%%%%%%%%%%%%%%%%%%%%%%%%%%%%%%%%%%%%
\section{Experiments}
\label{sec:exp}
\subsection{Setups}
\noindent
\textbf{Datasets.} We evaluate our model on four different multi-label image classification datasets. PASCAL VOC 2012 \cite{everingham2012pascal} has 5,717 training images and 5,823 images in the official validation set for testing, while PASCAL VOC 2007 contains a training set of 5,011 images and a test set of 4,952 images. MS-COCO \cite{lin2014microsoft} consists of 80 classes, with 82,081 training images and 40,137 testing images. NUSWIDE \cite{chua2009nus} has nearly 150K color images with various resolutions for training and 60.2K for testing, associated with 81 classes. The validation set is used for testing, whereas the training set is used to extract pseudo labels. During the training, we used these datasets without any ground truth. 

\smallskip
\noindent
\textbf{Implementation Details.} For initialization, to generate the pseudo labels based on our strategy, we use CLIP with ResNet-50$\times$64 as image encoder and keep the same CLIP Transformer \cite{vaswani2017attention,radford2019language} as the text encoder. During the training, for fair comparisons, all the models are trained using the same classification network architecture ResNet$-$101 \cite{he2016deep}, which is pre-trained on ImageNet \cite{deng2009imagenet} dataset. End-to-end training is used to update the parameters of the backbone and the classifier for 20 epochs. We train all four datasets using $10^{-5}$ learning rate. The batch size is chosen as $8$ for both VOC datasets, and $16$ for the COCO and NUS datasets. 

\smallskip
\noindent
\textbf{Pre-Training Setting.} As previously mentioned, our classification network is trained on unlabeled images without the use of any manual annotations during training; they are solely reserved for evaluation purposes. Therefore, we adapt CLIP using our global-local aggregation strategy to generate the pseudo labels for unlabeled data. We do not change or fine-tune the CLIP encoders or the prompt parameters, in which one fixed prompt is used, "a photo of the [class]", for all datasets.  To get the local similarity vectors, we split the input images into 3x3 snippet images to generate image embedding for each snippet, in addition to generating an embedding for the whole image to enhance the quality of the generated pseudo labels. All the unsupervised models are initialized and trained using our generated pseudo labels as initials for the unlabeled data. Additionally, CLIP is not exploited during the training or inference processes. 

\smallskip
\noindent
\textbf{Evaluation Metrics.}
For a fair comparison, we follow current works \cite{cole2021multi, kim2022large,abdelfattah2022g2netpl} that adopt the mean average precision (mAP), across the entire classes, as a metric for evaluation. We also measure the average precision (AP) per class to evaluate the class-wise improvement.
%
%%%%%%%%%%%%%%%%%%%%%%%%%%%%%%%%%%%%%%%%%%%%%%%%%%%%%%%%%%%
\begin{table*}[!ht]
    \centering{
    \caption{AP and mAP (in \%) of unsupervised methods on PASCAL VOC 2012 dataset for all classes. ALL methods trained by our proposed pseudo labels. Blue color represents the best results.}
      \label{tab:pascal-AP}
\begin{adjustbox}{width=1 \textwidth,totalheight={1.5cm}}
    \begin{tabular}%{lccccccccccccccccccccc}
            {
            %|
      >{}p{0.16\textwidth}
      >{\centering}p{0.065\textwidth}
     >{\centering}p{0.065\textwidth}
      >{\centering}p{0.065\textwidth}
     >{\centering}p{0.065\textwidth}
      >{\centering}p{0.065\textwidth}
         >{\centering}p{0.065\textwidth}
     >{\centering}p{0.065\textwidth}
      >{\centering}p{0.065\textwidth}
     >{\centering}p{0.065\textwidth}
      >{\centering}p{0.065\textwidth}
          >{\centering}p{0.065\textwidth}
     >{\centering}p{0.065\textwidth}
      >{\centering}p{0.065\textwidth}
     >{\centering}p{0.065\textwidth}
      >{\centering}p{0.065\textwidth}
         >{\centering}p{0.065\textwidth}
     >{\centering}p{0.065\textwidth}
      >{\centering}p{0.065\textwidth}
     >{\centering}p{0.065\textwidth}
      >{\centering}p{0.065\textwidth}  
      >{\centering\arraybackslash}p{0.065\textwidth}
    %   |
    }
    \hline
         Methods& aeroplane & bicycle & bird & boat & bottle & bus & car & cat & chair & cow & diningtable & dog & horse & motorbike & person & pottedplant & sheep & sofa & train & tvmonitor & mAP \\ \hline
     
                         %AN
       Naive AN \cite{kundu2020exploiting} & 98.9 & 92.3 & 96.2 & 86.6 & 70.0 & {94.7} & {83.6} & 97.6 & 73.3 & 80.4 &  {70.0} & 94.8 & 85.5 & 92.3 & 91.6 & 60.7 & 90.5 & 70.8 & 97.3 & {83.7} & 85.5\\
         %ANLS
        Szegedy \textit{et al.}~\cite{szegedy2016rethinking} & \textcolor{blue}{99.2} & {92.6} & {97.2} & {91.6} & 71.1 & 93.6 & 81.8 & 98.4 & {75.7} & {89.7} &69.4 & {96.2} & {86.9} & {93.9} & 92.5 & {65.6} & {93.2} & {72.3} & {98.2} & 77.3 & {86.8} \\ 
                %WAN
        Aodha \textit{et al.}\cite{mac2019presence} & 99.1 & 92.3 & 96.3 & 87.4 & {71.2} & 94.3 & 82.9 & 97.4 & 73.9 & 70.5 & 69.4 & 92.1 & 81.4 & 90.9 & 92.3 & 59.0 & 85.3 & 67.3 & 97.3 & 81.8 & 84.1 \\ 
        ROLE~\cite{cole2021multi} & 96.3 & 86.1 & 92.5 & 85.5 & 64.7 & 93.7 & 80.4 & 95.6 & 72.3 & 79.8 & 57.9 & 91.5 & 83.2 & 89.6 & {92.6} & 58.0 & 88.4 & 66.1 & 95.3 & 82.3 & 82.6\\ 
        % CLIP  \cite{radford2021learning} & 99.6 & 91.8 & \textbf{98.2} & \textbf{96.4} & 54.4 & \textbf{96.0} & 76.4 & \textbf{98.8} & 56.2 & \textbf{98.3} & \textbf{76.8} & \textbf{99.0} & \textbf{99.4} & \textbf{96.3} & 84.6 & 45.0 & \textbf{97.9} & 71.8 & \textbf{99.4} & 69.8 & 85.3\\ 
    
      \textcolor{blue}{CDUL (ours)}  & {99.0} &\textcolor{blue}{92.7} &\textcolor{blue}{97.7} &\textcolor{blue}{91.8} &\textcolor{blue}{72.5} &\textcolor{blue}{95.4} &\textcolor{blue}{84.7} &\textcolor{blue}{98.6} &\textcolor{blue}{76.4} &\textcolor{blue}{91.9} &\textcolor{blue}{73.2} &\textcolor{blue}{97.1} & \textcolor{blue}{92.0} & \textcolor{blue}{94.1} &\textcolor{blue}{93.0} &\textcolor{blue}{67.5} & \textcolor{blue}{94.2} &\textcolor{blue}{74.2} & \textcolor{blue}{97.7} &\textcolor{blue}{89.0} &\textcolor{blue}{88.6}\\ \hline
    \end{tabular}
    \end{adjustbox}
    }
\vspace{-0.5 cm}
\end{table*}

\subsection{Comparison with State-of-the-art Methods} 
\noindent
\textbf{Mean Average Precision mAP Results.} We report mAP (\%) results compared to the state-of-the-art models under different supervision levels in Table \ref{tab:supervised-level} on four different multi-label datasets. 
We compare our model to three different supervision levels. \textit{Fully supervised level} is used as a reference to the performance of the models using the fully labeled data \cite{cole2021multi,kim2022large}, upper part in Table \ref{tab:supervised-level}. At \textit{weakly supervised level}, we compare our model with \cite{kim2022large,abdelfattah2022g2netpl} methods that are trained using a single annotated label per each image following to \cite{kim2022large}. We also compare our model with another group of weakly supervised models that used a partial number of annotation labels (10\% per each image) for training such as \cite{Pu2022SARB,chen2022structured,ben2020asymmetric}. Finally, in the third group report in Table \ref{tab:supervised-level}, the \textit{unsupervised level}, all the methods  \cite{kundu2020exploiting,szegedy2016rethinking,mac2019presence,durand2019learning,cole2021multi}, including our method, are trained without any label annotation. 
We can observe that: \ding{182} Compared to fully supervised models,  we drastically removed the manual labeling costs without sacrificing performance compared to the fully supervised scenario.
\ding{183} Compared to weakly supervised models, our model can perform considerably better (mAP) comparable to those models without leveraging manually labeled data for the training set, which can be interpreted as meaning that our generated pseudo label that includes the high-quality fine-grained semantics based on our aggregator can help the classification network for training and get predictions that are competitive to those models that depend on partially annotated labels per image. Additionally, our model achieves better performance on the COCO and VOC 2007 datasets compared to the \cite{chen2022structured} method, which uses 10\% annotated label.
\ding{184} Compared to unsupervised models, our method outperforms the whole unsupervised models by a good margin on all datasets. The main reason is that our gradient-alignment optimization method can help the classification network to minimize the total loss based on the alternative updating methodology for the network parameters and the pseudo-label latent parameters. Our model can achieve +6.0\%, +4.4\%,and  +2.1\%, compared to Role \cite{cole2021multi} on VOC2012, VOC2007, and COCO,  respectively.
\textcolor{black}{Our method cannot be simply classified as weakly supervised models due to distinct input characteristics. Our approach utilizes CLIP to generate pseudo labels for all images, which often contain numerous unknown and incorrect labels (e.g., mAP using original CLIP is 65.3\% in COCO dataset). In contrast, weakly supervised models assumes that all provided partial labels are correct and can be trusted for training. This distinction is significant since our method tackles the joint training of the multi-label classification model and the refinement of incorrect pseudo labels, which is a key contribution of our work. Our method successfully increases the accuracy of CLIP-generated pseudo labels to 69.2\% mAP in COCO as reported in Table \ref{tab:supervised-level}.}
%in contrast to CLIP, which boasts a significantly larger number of parameters.}

\begin{figure}[!b]
  \centering
\vspace{-0.5 cm}
\includegraphics[scale=0.355]{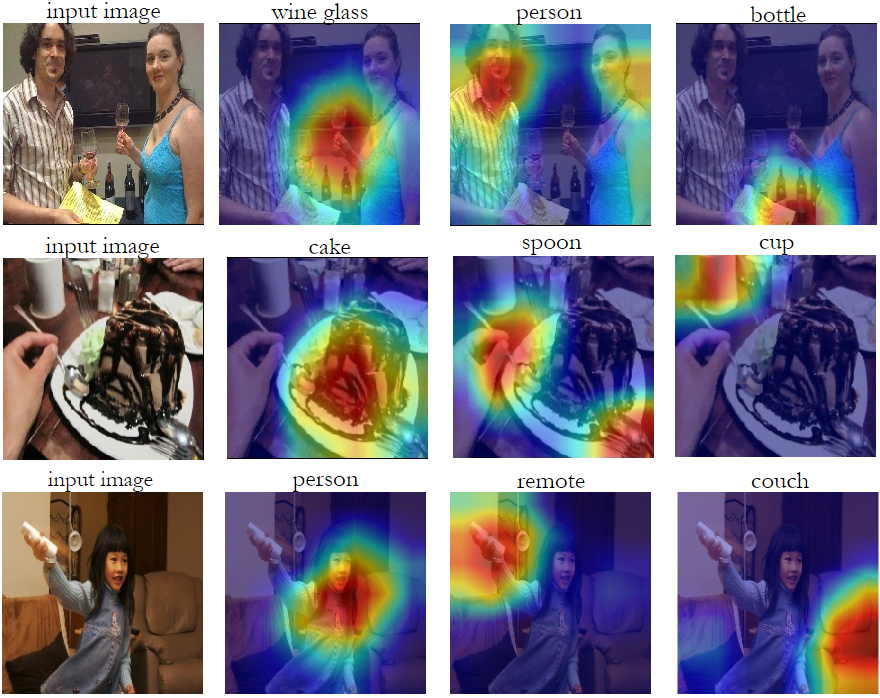}
  % \vspace{-0.25 cm}
  \caption{Class activation maps for several examples corresponding to highest confidences for three labels on COCO dataset. The highlighted area indicates where the model focused to classify the image. Best viewed in color. }
  \label{fig:cam-top3}
   \vspace{-0.5 cm}
\end{figure}

%
\smallskip
\noindent
{\bf Class-Wise AP Improvement.} 
Table \ref{tab:pascal-AP} reports the class-wise AP improvement for the unsupervised multi-label classification models on test sets of Pascal VOC 2012. All the methods start training based on our proposed global-local pseudo labels. Our method can improve performance in most classes in VOC 2012 dataset. We observe that although all the methods start training based on our proposed global-local pseudo labels, our model outperforms in most of the classes, especially the classes that have a small size, such as \textit{"potted plant"}, \textit{"book"}, \textit{"cup"}, and \textit{"wine glass"} which can be interpreted that gradient-alignment can help our model to capture more information during the training.
We also demonstrate the Class Activation Maps (CAM) for some images in the COCO dataset, as shown in Figure \ref{fig:cam-top3}. Our model can classify the \textit{"bottle"} and \textit{"wine glass"} in the first row, \textit{cup} in the second row, and \textit{"remote"} in the third row.
%%%%%%%%%%%%%%%%%%%%%%%%%%%%%%%%%%%%%%%%%%
%############################################
\subsection{Ablation Study}
\begin{figure}[!t]
  \centering
\includegraphics[scale=0.23]{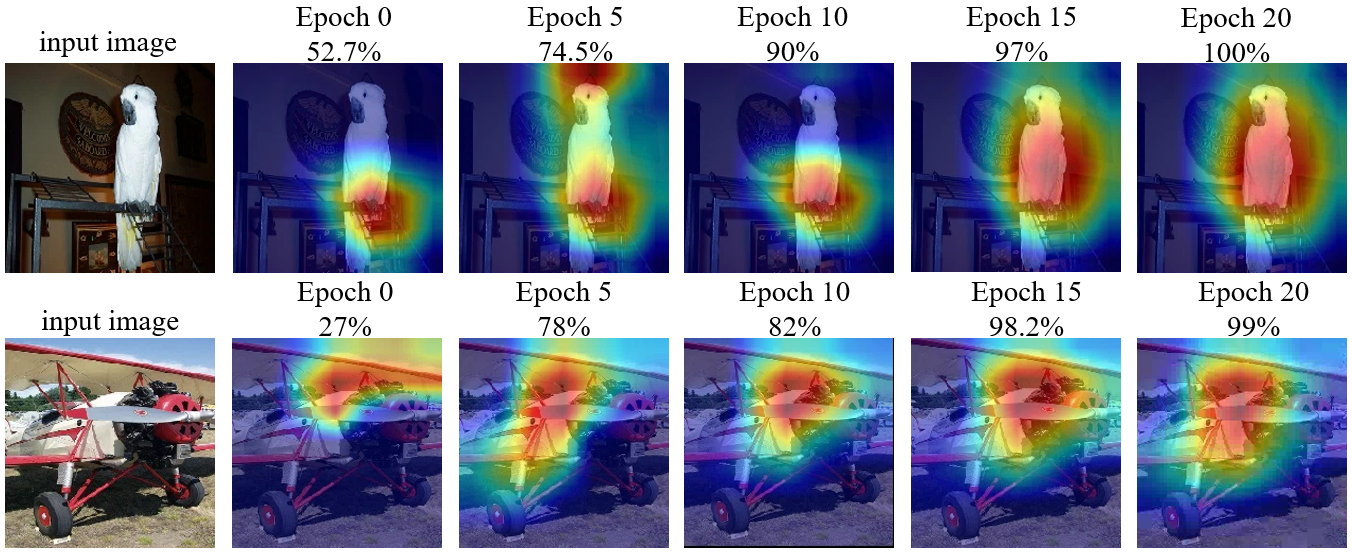}
  \caption{CAM visualization for the classification task on Pascal2012 dataset. CAM shows that the improvement of the classification during the epoch. Best viewed in color.}
  \label{fig:cam}
  \vspace{-0.3 cm}
\end{figure}

\noindent
{\bf Quality of Initial Pseudo Labels.}
\label{subsec:Quality}
To study the quality of the pseudo-labels, we measure the mAP for the CLIP-pseudo labels based on the global and local alignments using our proposed aggregation strategy (ours) as reported in Table \ref{tab:quality-pseudo-labels}. We also report different aggregation strategies such as \ding{182} average (avg): by getting the average between all the local and global similarity vectors, \ding{183} Maximum (max): to get the maximum similarity score per each class among all the local and global similarity vectors. As reported in Table \ref{tab:quality-pseudo-labels}, we observe that the quality of pseudo label for the global alignment achieves mAP less than any strategy depending on local alignment. We also observe that our aggregation strategy generates the highest quality pseudo labels compared to its global counterparts +5\%, 7.4\%, and 1.9\% on Pascal VOC 2012, COCO, and NUS datasets, respectively. Consequently, our aggregation strategy can retain the most fine-grained semantics in the input image. We also prove that the good quality of the pseudo label helps the classification network to learn during the epochs and the classification performance is improved by demonstrating the CAM visualization for samples of images from Pascal VOC 2012. As shown in Figure \ref{fig:cam}, during the epochs, the classification network can learn the correct prediction and mAP is increased.
%
%%%%%%%%%%%%%%%%%%%%%%%%%%%%%%%%%%%%%%%%%%
\begin{table}
    \centering
    \caption{Ablation study of quality of pseudo labels on the training set in three different datasets using ResNet-50 $\times$ 64.}
    \label{tab:quality-pseudo-labels}
    \begin{adjustbox}{width=0.4 \textwidth}
    \begin{tabular}%{|l|l|l|l|l|}
     {
     >{}p{0.1\textwidth}|
      >{\centering}p{0.09\textwidth}|
     >{}p{0.06\textwidth}
      >{\centering}p{0.06\textwidth}
     % >{\centering}p{0.07\textwidth}
      >{\centering\arraybackslash}p{0.06\textwidth}
    }
    \hline 
    \multirow{3}{*}{Datasets} & \multirow{3}{1.0cm}{global alignment}  &\multicolumn{3}{c}{global-local alignment} \\ \cline{3-5}
       ~ & ~  &\multicolumn{3}{c}{Aggregator}   \\ \cline{3-5}
        ~ & ~ & avg & max & ours  \\ \hline
        VOC 2012 & 85.3 & 88.5 & 89.5 & {90.3}  \\ \hline
        COCO & 65.4 & 70.0 & 71.6 & {72.8}   \\ \hline
        NUS & 41.2 & 41.8  & 42.3 & {43.1}  \\ \hline
    \end{tabular}
    \end{adjustbox}
        \vspace{-0.3 cm}
\end{table}
% \multicolumn{3}{*}{Aggregator}
%%%%%%%%%%%%%%%%%%%%%%%%%%%%%%%%%%%%%%%%%%

%%%%%%%%%%%%%%%%%%%%%%%%%%%%%%%%%%%%%%%%%%
% \vspace{-0.5 cm}
%############################################
%############################################
% \vspace{-0.5 cm}
\begin{figure}[!t]
  \centering
  	\includegraphics[scale=0.54]{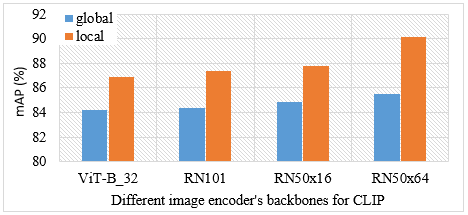}\\
  % \vspace{-0.25 cm}
  \caption{Quality of pseudo labels using different backbones for CLIP's image encoder}
  \label{fig:backbones}
  \vspace{-0.5 cm}
\end{figure}
%#################################
\smallskip
\noindent
{\bf Different Backbones. }
In our study, we evaluate the quality of pseudo labels generated using different depths of image encoders in CLIP, namely ViT-B-32, ResNet-101, ResNet-50 $\times$ 16, and ResNet-50 $\times$ 64 following the EfficientNet-style model scaling \cite{radford2021learning}. We employ CLIP with different backbones during the initialization time to generate the pseudo labels on the training set of Pascal VoC 2007 dataset. The results in Figure \ref{fig:backbones} show that the quality of the generated pseudo labels consistently improves with different backbones. Furthermore, we observed that the quality of pseudo labels improves significantly with the use of local alignment, achieving up to a 2.7\% improvement for ViT-B-32, 3\% for ResNet-101, 2.9\% for ResNet-50 $\times$ 16, and 4.6\% for ResNet-50 $\times$ 64, compared to their global alignment counterparts. Since we use this backbone only once at the initialization to generate the pseudo labels, no computational cost will be used at the testing time. Additionally, the generated pseudo labels are used as initials for all unsupervised multi-label models. As reported in Table \ref{tab:our-backbone}, the performance of our model is improved with different backbones, achieving up to (+2\%) increase using ResNet-50 $\times$ 64 compared to its counterparts. The improvement is reasonably related to the quality of pseudo labels used during the training Figure \ref{fig:backbones}.

\smallskip
\noindent
% {\bf {Different Loss Functions for Pseudo Labels}.}
% {We investigate various approaches to update the pseudo labels, including binary cross entropy (BCE) and momentum curriculum learning (MCL) with $\psi(y_u)$ kept in all cases, as reported in Table \ref{tab:different_loss}. Our approach using KL divergence is compared with these alternatives by training ResNet-50 for three different scenarios on two different datasets. Our approach outperforms the other methods because, in the context of multi-label classification with pseudo-labels, the KL loss function can measure the difference between the predicted probability distribution and the modified target probability distribution constructed from the pseudo-labels, which helps during model training.}
% , as explained in section \ref{subsec:GANT}.

\smallskip
\noindent
{\bf CLIP Performance on Test Set.} {Table~\ref{tab:GLA} presents a comparison of our ResNet-50-based model's performance with on-the-shelf CLIP with ResNet-50 combined with our global-local alignment strategy (CLIP-GLA). Our model surpasses CLIP-GLA in both mAP.  Moreover, our model uses a smaller number of parameters during inference. This is due to the fact that in our framework, CLIP is used exclusively to initialize the pseudo labels during the initialization phase, while a smaller backbone is utilized during the training and testing phases, as illustrated in Fig. \ref{fig:clip-main-framework}. Thus, our approach is more cost-effective during both the training and inference phases.
}
%textcolor{blue}{It is worth noting that the number of parameters of CLIP is 623M, whereas only 25M parameters if ResNet-50 is utilized as a backbone or 44M parameters if ResNet-101 is employed. Therefore, CLIP is employed solely for initializing the pseudo labels only at initialization time, and we opt for a smaller backbone during the training and testing phases, as depicted in Fig. \ref{fig:clip-main-framework}. }
%###############################
\begin{table}[!t] 
    \centering
    \caption{Ablation study when initialized with various pseudo labels based on different CLIP's image encoder backbones. mAP results on Pascal 2007 dataset}
        \label{tab:our-backbone}
        \small
    \begin{tabular}{lcccc}
    \hline
         ViT-B-32 & ResNet-101 & ResNet-50x16 & ResNet-50x64  \\ \hline
         86.7 & 86.8 & 86.9 & 89.0  \\ \hline
        % ~ & ~ & ~ & ~ & ~ \\ \hline
        % ~ & ~ & ~ & ~ & ~ \\ \hline
        % ~ & ~ & ~ & ~ & ~ \\ \hline
    \end{tabular}
      \vspace{-0.1 cm}
\end{table}
%%%%%%%%%%%%%%%%%%%%%%%%%%%%%%%%
%###############################
% \begin{table}[!t] 
%     \centering
%     \caption{Ablation study for updating the pseudo labels.}
%         \label{tab:different_loss}
%         \small
%     \begin{tabular}%{lcccc}
%          {
%           >{}p{0.08\textwidth}
%       >{\centering}p{0.08\textwidth}
%      >{\centering}p{0.11\textwidth}
%       % >{\centering}p{0.05\textwidth}
%      % >{\centering}p{0.07\textwidth}
%       >{\centering\arraybackslash}p{0.09\textwidth}
%     }
%     \hline
%          Datasets &BCE & MCL \cite{abdelfattah2022plmcl} & KL(Ours) \\ \hline
%          COCO & 67.9 & 68.2 & 68.4   \\ 
%          Pasca2012 & 87.0 & 87.5 & 87.8   \\ \hline
%         % ~ & ~ & ~ & ~ & ~ \\ \hline
%         % ~ & ~ & ~ & ~ & ~ \\ \hline
%         % ~ & ~ & ~ & ~ & ~ \\ \hline
%     \end{tabular}
%       \vspace{-0.1 cm}
% \end{table}
%%%%%%%%%%%%%%%%%%%%%%%%%%%%%%%%
%###############################
\begin{table}[!t] 
    \centering
    \caption{Ablation study to evaluate CLIP-GLA's performance.}
        \label{tab:GLA}
        \small
    \begin{tabular}{l|ccc|c}
    \hline
         Models &Pascal2012 & COCO & NUS & \# param. (M) \\ \hline
         CLIP-GLA & 84.7 & 63.6 & 38.9 & 102  \\ \hline
         CDUL (ours) & 86.4 & 67.1 & 41.9 & 25 \\ \hline
        % ~ & ~ & ~ & ~ & ~ \\ \hline
        % ~ & ~ & ~ & ~ & ~ \\ \hline
    \end{tabular}
      \vspace{-0.4 cm}
\end{table}
%%%%%%%%%%%%%%%%%%%%%%%%%%%%%%%%

\vspace{-2mm}
\noindent 
{\bf Effectiveness of Other Parameters.} We also study the impact of removing the Gaussian distribution module in the gradient-alignment training, the performance is dropped by 0.5\%, and 0.4\% on VoC 2012, and COCO datasets, respectively, as compared to our model (Table \ref{tab:supervised-level}). Additionally, we study applying the hard pseudo labels instead of soft pseudo labels; the performance is reduced by 0.9\%, and 1.2\% on VoC 2012, and COCO datasets, respectively, as compared to our model (Table \ref{tab:supervised-level}).

\section{Conclusions}
\label{sec:con}
\vspace{-0.15 cm}
In this paper, we propose a new method for unsupervised multi-label image classification tasks without using human annotation. Our key innovation is to modify the vision-language pre-train model to generate the soft pseudo labels, which can help training the classification network. To inject the fine-grained semantics in the generated pseudo labels, we proposed a new aggregator that combines the local(global) similarity vectors between image snippets(whole image) and text embedding. Finally, we use the generated pseudo label to train the network classification based on the gradient-alignment to get multi-label classification prediction without any annotation. Extensive experiments show that our method outperforms state-of-the-art unsupervised methods.

% \smallskip
% \noindent
% {\bf Discussions.} Collecting massive amounts of data with completely labeled categories is challenging in multi-label classification task, making an unsupervised setup vital to save the cost of annotation. Although we have comparable results compared to weakly supervised methods, there is still a lot of room for improvement in finding the relationship between the labels, which will be helpful in different computer vision applications.
%%%%%%%%% REFERENCES
\clearpage

{\small
\bibliographystyle{ieee_fullname}
\bibliography{egbib}
}
\clearpage
\begin{spacing}{1.45} 
\section{Supplementary Material}
\subsection{Initialization:}
\label{sec:1}
%\textbf{Cutting the images:} The input images from various datasets are resized to square images measuring $672 \times 672$ pixels. Subsequently, each image is partitioned into a grid of 3 rows and 3 columns, resulting in snippets at $224 \times 224$ pixels each.

\textbf{Global-Local Aggregator:} \\
We used the off-shelf CLIP to get similarity scores for both the entire input image (referred to as "global") and and each individual snippet within the image (referred to as "local").  Subsequently, we employed two different aggregation approaches to get $S^{final}$: (I) aggregation based on maximum ($\lambda=1$): by getting the maximum similarity score per each class among global and max-min local similarity vectors, and (II) aggregation based on average ($\lambda=0$): by averaging between the global similarity scores and max-min of local similarity scores for each class. $S^{final}$ is described as follow;
\begin{equation*}
S^{final} = \frac{1}{2} \left(S^{global} + S^{aggregate} + \lambda |S^{global} - S^{aggregate}|\right),
% \vspace{-0.25 cm}
\end{equation*}
where $\lambda$ is the smoothing hyper-parameter changes between the aggregation based on maximum to aggregation based on average across the global and min-max local similarity scores. 
% The mAPs of the quality of pseudo labels based on two datasets are reported in Table \ref{tab:quality-pseudo-labels1}. 
%%%%%%%%%%%%%%%%%%%%%%%%%%%%%%%%%%%%%%%%%%
% \begin{table}[!h] 
%     \vspace{0.3 cm}
%     \centering
%     \caption{Ablation study of quality of pseudo labels on the training set in two different datasets using ResNet-50 $\times$ 64.}
%     \label{tab:quality-pseudo-labels1}
%     \begin{adjustbox}{width=0.4 \textwidth}
%     \begin{tabular}%{|l|l|l|l|l|}
%      {
%      >{}p{0.1\textwidth}|
%       >{\centering}p{0.08\textwidth}|
%      >{\centering}p{0.08\textwidth}
%      % >{\centering}p{0.07\textwidth}
%       >{\centering\arraybackslash}p{0.08\textwidth}
%     }
%     \hline 
%     \multirow{2}{*}{Datasets} & \multirow{2}{1.0cm}{global alignment}  &\multicolumn{2}{c}{global-local alignment} \\ 
%        ~ & ~  &\multicolumn{2}{c}{Aggregator}   \\ \cline{3-4}
%         ~ & ~  & case (I) & case (II)  \\ \hline
%          % ~ & ~ & avg & max &   \\ \hline
%         VOC 2012  & 88.5 & 89.5 & {90.3}  \\ \hline
%         VOC 2007  &  &90.1  &90.9  \\ \hline
%         %COCO & 65.4 & 70.0 & 71.6 & {72.8}   \\ \hline
%         %NUS & 41.2 & 41.8  & 42.3 & {43.1}  \\ \hline
%     \end{tabular}
%     \end{adjustbox}
%         % \vspace{0.3 cm}
% \end{table}
% \multicolumn{3}{*}{Aggregator}
%%%%%%%%%%%%%%%%%%%%%%%%%%%%%%%%%%%%%%%%%%
\\
% We additionally provide the mAPs for two datasets across various $\zeta$ values in range from 0.5 to 0.  As shown in Figure \ref{tab:table2}, across different $\zeta$ values, the mAPs of the quality of pseudo labels outperform their corresponding global alignment for each dataset with significant margins \textcolor{red}{as shown in Figure}. We reported the highest values for mAP based on $\zeta = 0$ in Tables 3 and 6.
%###############################
% \begin{table}[!h] 
%     \centering
%     \caption{Ablation study on different values of $\zeta$ for case (III). mAP results on Pascal 2012 dataset \textcolor{red}{we can use column figure for global and $\zeta$ instead of table}}
%         \label{tab:table2}
%         \small
%     \begin{tabular}{l|cc}
%     \hline
%          ~ &\multicolumn{2}{c}{mAP}   \\ \cline{2-3}
%          $\zeta$ & pascal 2012 & coco  \\ \hline
%           0.4&  & \\ \hline
%           0.3& &  \\ \hline
%           0.2&  & \\ \hline
%           0.1& &  \\ \hline
%           0& 90.3(> 5.0\%) & 72.8 (> 7.4\%) \\ \hline
%     \end{tabular}
%       \vspace{-0.1 cm}
% \end{table}

\subsection{During the training:}
We trained the network using Kullback-Leibler (KL) loss function. Then fix the predicted labels to update the latent parameters of pseudo labels using equation (7), and Gaussian distribution with the mean at 0.5, which is given by:
\vspace{-0.25 cm}
\begin{align}
    \psi ([{y}_{u}]_i) = \frac{c1}{\sigma \sqrt{2\pi}} e^{- \frac{1}{2} \left(\frac{[{y}_{u}]_i-0.5}{\sigma}\right)^2} + c2 \nonumber
    \label{eq:psi}
    \vspace{-0.25 cm}
\end{align}

where $c1$ and $\sigma$ are the hyperparameters and $c2$ is a constant that ensures the Gaussian function $\psi ([{y}_{u}]_i)$ close to zero when $[{y}_{u}]_i$ has high confidence score with values at 0 and 1. At epoch 0, the latent parameters are initialized with pseudo labels obtained during the initialization phase via a local-global aggregator. We used warm-up until epoch 3 without updating the pseudo labels. Starting from epoch 4, the network parameters and the latent parameters of pseudo labels are updated alternatively and reported the results at epoch 20. We initialized the latent parameters with pseudo labels aggregated in different cases, as discussed in section \ref{sec:1}. For example, the pseudo labels are initialized with aggregated scores at $\lambda=1$ in Table 1. The $\zeta$ values range from $0$ to $0.4$, where the global-local aggregated pseudo labels can achieve mAPs higher than the global pseudo labels. 

%The highest mAP of quality of pseudo labels is achieved at $\zeta = 0$.} As future work, we will study the comparison between different local-global settings. 

%where $c1 = 2$, and $\sigma = 0.15$ and \textcolor{red}{we used these hyper parameters for all the datasets.} 

% %###############################
% \begin{table}[!h] 
%     \centering
%     \caption{mAP results on \textcolor{red}{four} datasets}
%         \label{tab:our-backbone}
%         \small
%     \begin{tabular}{lcccc}
%     \hline
%          dataset  & mAP & weights  \\ \hline
%          pascal 2012 &  &  \\ \hline
%          pascal 2007 &  &  \\ \hline
%          coco &  &  \\ \hline
%         % NUS &  &  \\ \hline
%     \end{tabular}
%       \vspace{-0.1 cm}
% \end{table}
% %%%%%%%%%%%%%%%%%%%%%%%%%%%%%%%%
\subsection{Testing Phase:} During the test phase, we only used the network to test the input image, where the network takes an entire image as input rather than snippets.
\section{Evaluation Metrics} \label{sec: evaluation}
This section introduces the metrics used to evaluate
the performance of the network for multi-label image classification. We assume that each image is assigned with the estimated label vector $y_o$, whose entries are soft pseudo labels from the global-local aggregator. During testing, each image is associated with the fully labeled ground truth $y_g$, whose entries can be $1$ or $0$, representing observed positive or observed negative labels, respectively. 

%\subsection{Average Precision (AP) and Mean Average Precision (mAP)} 
The mean average precision~(mAP) is applied to evaluate the performance of different approaches for multi-label classification in our paper, similar to \cite{cole2021multi,huynh2020interactive}.  
We measure the average precision (AP) for each class to calculate
the mAP across all $L$ classes as following:
\begin{align}
mAP = \frac{1}{L} \sum_{\ell=1}^L AP_{\ell}.
\end{align}

\end{spacing}
\end{document}